\documentclass[runningheads]{llncs}
\usepackage{graphicx}
\usepackage{amsmath,amssymb}

\usepackage{color}
\usepackage{float}
\usepackage{caption}
\usepackage{multirow}
\usepackage{hyperref}
\usepackage{rotating}
\usepackage{makecell}

\def\etal{et al.}

\begin{document}

\title{Class-Agnostic Counting} 
\titlerunning{Class-Agnostic Counting} 

\author{Erika Lu \and Weidi Xie \and Andrew Zisserman}

%

\authorrunning{E. Lu et al.} 

\institute{Visual Geometry Group, University of Oxford\\
\email{\{erika,weidi,az\}@robots.ox.ac.uk}}
\maketitle



\begin{abstract}
Nearly all existing counting methods are designed for a specific object class. 
Our work, however, aims to create a counting model able to count any class of object.
To achieve this goal, 
we formulate counting as a matching problem, 
enabling us to exploit the image self\-similarity property 
that naturally exists in object counting problems.

We make the following three contributions:
\emph{first}, 
a Generic  Matching Network (GMN) architecture that can potentially count any object in a class-agnostic manner; 
\emph{second}, by reformulating the counting problem as one of matching objects, 
we can take advantage of the abundance of video data labeled for tracking,
which contains natural repetitions suitable for training a counting model.
Such data enables us to train the GMN.
\emph{Third}, to customize the GMN  to different user requirements,
an adapter module is used to specialize the  model with minimal effort, 
i.e.\ using a few labeled examples, and adapting only a small fraction of the trained parameters.
This is a form of few-shot learning,
which is practical for domains where labels are limited due to requiring expert knowledge (e.g.\ microbiology).

We demonstrate the flexibility of our method on a diverse set of existing counting benchmarks:
specifically cells, cars, and human crowds.
The  model achieves competitive performance on cell and crowd counting datasets,
and surpasses the state-of-the-art on the car dataset using only three  training images.
When training on the entire dataset, the proposed method outperforms all previous methods by a large margin.
\end{abstract}
\section{Introduction}
\label{sec:intro}
The objective of this paper is to count objects of interest in an image.
In the literature, object counting methods are generally cast into two categories:
detection-based counting~\cite{Barinova10,Desai09,Hsieh17} 
or regression-based counting~\cite{Arteta14,Arteta16,Cho1999,Kong06,Lempitsky10b,Marana97,Xie15}.
The former relies on a visual object detector that can localize object instances in an image; this
method, however, requires training individual detectors for different objects, 
and the detection problem remains challenging if only a small number of annotations are given.
The latter avoids solving the hard detection problem --
instead, 
methods are designed to learn either a mapping from global image features to a scalar~(number of objects),
or a mapping from dense image features to a density map, achieving 
better results on counting overlapping instances. However, previous methods for both categories of method (detection, regression) have only developed algorithms that can count a particular class of objects (e.g.\ cars, cells, penguins, people).

\begin{figure}[t]
\centering
\includegraphics[width=\textwidth]{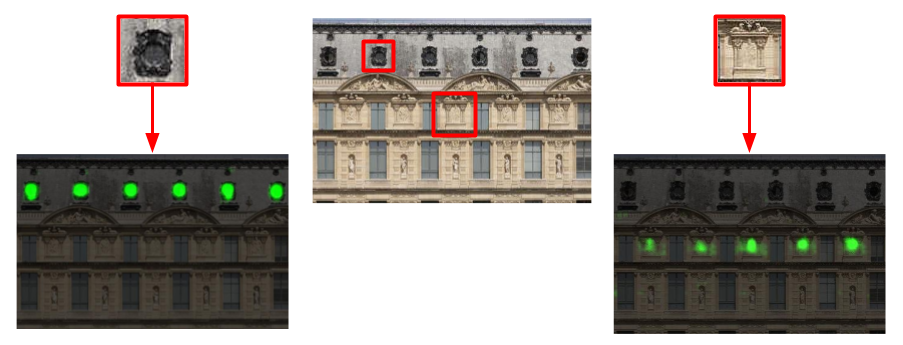}
\caption{
The model (trained on tracking data) can count an  object, e.g.\  windows or columns,  specified  
as an exemplar patch (in red), without additional training. The heat maps indicate the localizations of the repeated objects.  This image is unseen during training.}
\label{fig:teaser}
\end{figure}

The objective of this paper is a class-agnostic counting network -- one that is able to flexibly count object instances in an 
image by, for example, simply specifying an exemplar patch of interest as illustrated in Figure~\ref{fig:teaser}. To 
achieve this, we build on a property of images that has been largely ignored explicitly in previous counting 
approaches -- that
of  \emph{image self-similarity}. 
At a simplistic level, 
an image is deemed self-similar if patches repeat to some approximation -- for example 
if patches can be represented by other patches in the same image.
Self-similarity  has underpinned applications for  many vision tasks, 
ranging from texture synthesis~\cite{Efros99}, to 
image denoising~\cite{Buades2005}, to
super-resolution~\cite{Glasner2009}.

Giving the observation of self-similarity, image counting can be recast as an image {\em matching} problem -- counting
instances is performed by matching (self-similar patches) within the same image. 
To this end we develop a {\em Generic Matching Network} (GMN) that learns a discriminative classifier
to match instances of the exemplar. 
Furthermore, since matching variations of an object instance  within an image is similar to  matching variations of
an object instance between images, 
we can take advantage of the abundance of video data labeled for tracking
which contains natural repetitions, to train the GMN.
This observation, that matching within an image can be thought of as tracking within an image, was previously
made by Leung and Malik~\cite{Leung96} for the case of repeated elements in an image.

Beyond generic counting, there is often a need to {\em specialize} matching to more restrictive or general requirements.
For example, to count only red cars (rather than all cars) or to count cars at all orientations (which goes beyond
simple similarity measures such as squared sum of differences), extending the
intra-class variation for the object category of interest~\cite{Han15,Gregory2015}.
To this end, we include an  adaptor module that enables 
fast domain adaptation~\cite{Rebuffi18} and few-shot learning~\cite{Sung2017,Vinyals16},
through the training of a small number of tunable parameters, using very few annotated data.

In the following sections, 
we begin by detailing the design and training procedure of the GMN  in \S~\ref{sec:methods}, 
and demonstrate its  capabilities on a set of example counting tasks.
In~\S~\ref{sec:exp}, we adapt the GMN to specialize on
several counting benchmark datasets, 
including the VGG synthetic cells, HeLa cells, and cars captured by drones.
During adaptation,
only a small number of parameters (3\% of the  network size) are added and trained on the target domain.
Using a very small number of training samples~(as few as $3$ images for the car dataset), 
the results achieved are either comparable to, or surpass the current state-of-the-art methods by a large margin.
In~\S~\ref{sec:exp2}, 
we further extend the counting-by-matching idea to a more challenging scenario: 
Shanghaitech crowd counting, 
and demonstrate promising results by matching image statistics
on scenes where accurate instance-level localization is unobtainable.
\section{Method}
\label{sec:methods}
In this paper, we consider the problem of instance counting, 
where the objects to be counted in a single query are from the same category,
such as the windows on a building, cars in a parking lot, or cells of a certain type. 

To exploit the \emph{self-similarity} property,
the counting problem is reformalized as localizing  and counting  ``repeated'' instances by~\textit{matching}.
We propose a novel architecture -- GMN, and a counting approach
which requires learning a comparison scheme for two given objects (patches) in a metric space. 
The structure of the model naturally accommodates class-agnostic counting, as it 
learns to search for repetitions of an exemplar patch containing the desired instance.
Note that, the concept of~\emph{repetition} is defined in a very broad sense; 
in the following experiments,
we show that objects with various shapes, overlaps, and complicated appearance changes can still be treated as ``repeated'' instances.

\begin{figure}[!htb]
\begin{center}
\rule{0pt}{1ex}
\includegraphics[width=\textwidth]{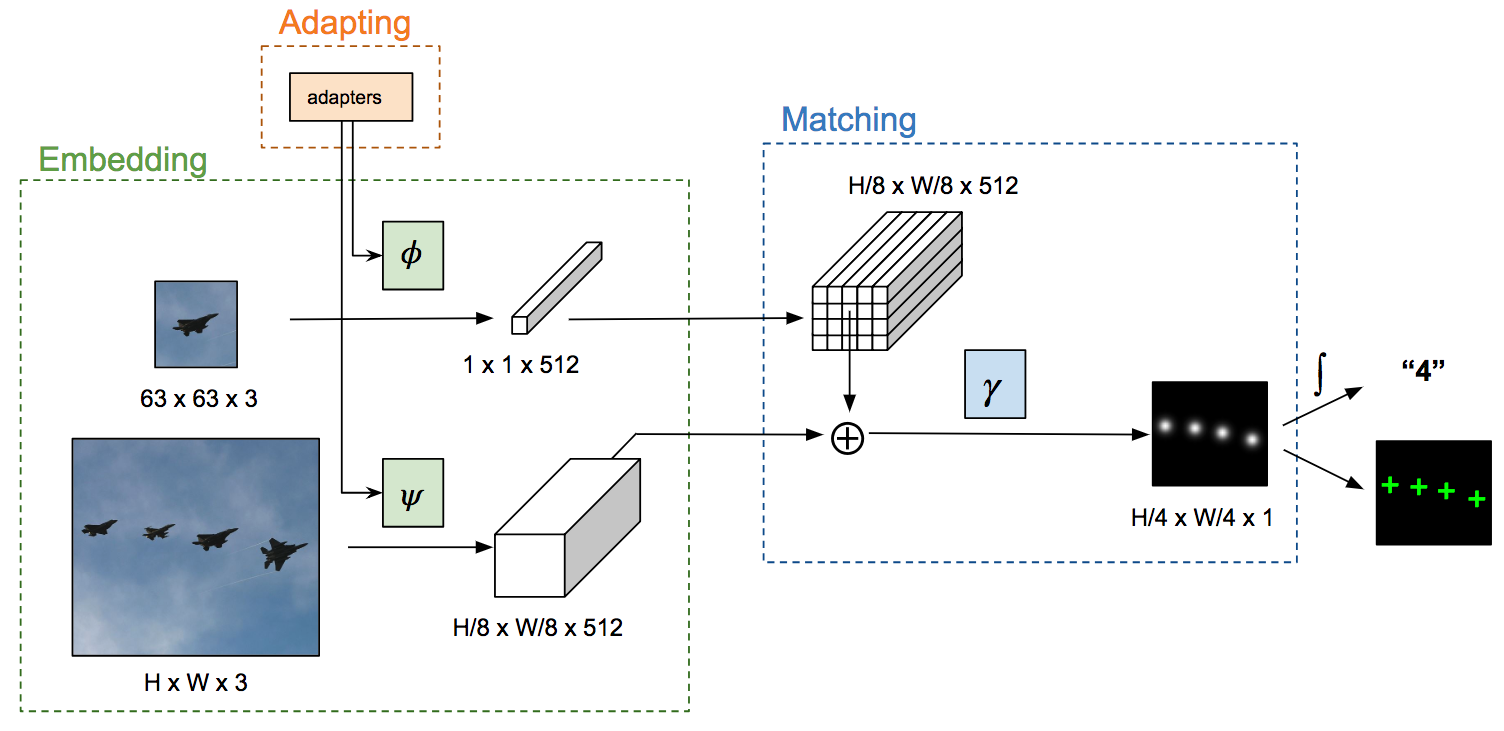}
\caption{The GMN architecture consists of three modules: embedding, matching, and adapting. 
The final count and detections are obtained by taking the integral and local maximums, 
respectively, over the output heatmap. 
The integral-based method is used for counting instances with significant overlap, 
while the local max is used where objects are well-separated
and the positional information is of interest for further processing.
The $\bigoplus$ represents channel-wise concatenation.}
\label{fig:model}
\end{center}
\end{figure}

The entire GMN consists of three modules, namely, \emph{embedding}, \emph{matching}, and \emph{adapting}, as illustrated in Figure~\ref{fig:model}.
In the \emph{embedding} module, 
a two-stream network is used to encode the exemplar image patch and the full-resolution image 
into a feature vector and dense feature map, respectively.
In  the \emph{matching} module, 
we learn a discriminative classifier
to densely match the exemplar patch to instances in the image.
Such learning overcomes within image variations such as 
illumination changes, small rotations, etc.
The object locations and final count can then be acquired by simply taking the \emph{local maximums} 
or \emph{integration} over the output similarity maps, respectively.
Empirically, 
integral-based counting shows better performance in scenarios where instances have significant overlap,
while local max counting is preferred where objects are well-separated, 
and the positional information is of interest for further processing (e.g.\ seeds for segmentation).

To avoid the time-consuming collection of annotations for counting
data, we use the observation that repetitions occur naturally in
videos, as objects are seen under varying viewing conditions from
frame to frame.  Consequently, we can train the generic matching network with the
extensive training data available for tracking~(specifically the
ILSVRC video dataset for object detection~\cite{Russakovsky15}).  In
total, the dataset contains nearly $4500$ videos and over $1M$
annotated frames.

Given a trained matching model, several factors can prevent it
from generalizing perfectly onto the target domain: for instance, the
image statistics can be very different from the training set (e.g.\
natural images vs.\ microscopy images), or the user requirements can
be different (e.g.\ counting cars vs.\ counting only red cars).  Efficient domain adaptation requires a module that
can change the network activations with minimal effort (that is, minimal
number of trainable parameters and very few training data). Thus,
for the \emph{adapting} stage, we incorporate residual adapter modules~\cite{Rebuffi18} to specialize the
GMN to such needs.
Adapting to the new counting task then merely involves freezing the
majority of parameters in the generic matching network, 
and training the adapters~(a small number of extra parameters) on the
target domain with only a few labeled examples.

\subsection{Embedding}
In this module, 
a two-stream network is defined for transforming raw RGB images into high-level feature encodings.
The two streams are parametrized by separate functions for higher representation capacity: 
\[
v = \phi(z; \theta_1) ~~~~~~~~~~
f = \psi(x; \theta_2)
\]
In detail, the function $\phi$  transforms an exemplar image patch $z \in R^{63 \times 63 \times 3}$ 
to a feature vector $v \in R^{1 \times 1 \times 512}$,
and $\psi$  maps the full image $x \in R^{H \times W \times 3}$ to a feature map $f \in R^{H/8 \times W/8 \times 512}$. 
Both the vector $v$ and feature maps $f$ are L2 normalized along the feature dimensions.
In practice, our choices for $\phi(\cdot;\theta_1)$ and $\psi(\cdot;\theta_2)$ are ResNet-50 networks~\cite{He2016} truncated after the final conv3\textunderscore x layer. 
The resulting feature map from the image patch is globally max-pooled into the feature vector $v$. 

\subsection{Matching}
The relations between the resulting feature vector and maps are modeled 
by a trainable function $\gamma(\cdot;\theta_3)$ that takes the concatenation of $v$ and $f$ as input, 
and outputs a similarity heat map, as shown in Figure~\ref{fig:model}.
Before concatenation, $v$ is broadcast to match the size of the feature maps to accommodate the fully convolutional feature, which allows for efficient modeling of the relations between the exemplar object and all other objects in the image.
The similarity $Sim$ is given by
\begin{align*}
Sim = \gamma([broadcast(v): f]; \theta_3)
\end{align*}
\noindent 
where ``$:$'' refers to concatenation, 
and $\gamma(\cdot;\theta_3)$ is parametrized by one $3\times3$ convolutional layer 
and one $3\times3$ convolutional transpose layer with stride 2~(for upsampling).
\subsection{Training Generic Matching Networks}
The  generic matching network~(consisting of embedding and matching modules) is trained 
on the ILSVRC video dataset.
The ground truth label is a Gaussian placed at each instance location, 
multiplied by a scaling factor of $100$, and a weighted MSE~(Mean Squared Error) loss is used.
Regressing a Gaussian allows the final count to be obtained by simply summing 
over the output similarity map, which in this sense doubles as a density map.

During training, 
the exemplar images are re-scaled to size $63\times63$ pixels,
with the object of interest centered to fit the patch,
and the larger $255\times255$ search image is taken as a crop centered around the scaled object
(architecture details can be found in Table~\ref{net:tracknet}.
More precisely, 
we always scale the search image according to the bounding box~(w,h) of the exemplar objects, 
where the scale factor is obtained by solving $s \times h \times w = 63^2$.
The input data is augmented with horizontal flips and small ($<25^\circ$) rotations and zooms,
and we sample both positive and negative pairs.
In all subsequent experiments, the network has been pre-trained as described here. 
\begin{table}[H]
\renewcommand\arraystretch{1.4}
\begin{center}{\scalebox{0.87}{
\begin{tabular}{|c|p{3.5cm}<{\centering}|p{3.5cm}<{\centering}|p{3.5cm}<{\centering}|}
\hline
Module 
& Exemplar Patch ($N \times 63 \times 63 \times 3$) 
& Image to Count ($N \times 255 \times 255 \times 3$)
& Output Size \\
\hline
\multirow{12}{*}{\emph{Embedding}}
& \multicolumn{1}{c|}{conv, $7\times7$, $64$, stride $2$} 
& \multicolumn{1}{c|}{conv, $7\times7$, $64$, stride $2$} 
& $N \times 32 \times 32 \times 64$     $N \times 128 \times 128 \times 64$    \\
\cline{2-4}
& \multicolumn{1}{c|}{max\;pool, $3\times3$, stride $2$}
& \multicolumn{1}{c|}{max\;pool, $3\times3$, stride $2$} 
& \\
& \multicolumn{1}{c|}
{$\begin{bmatrix} {\rm conv}, 1\times 1, 64 \\ {\rm conv}, 3\times 3, 64 \\ {\rm conv}, 1\times 1, 256 \end{bmatrix} \times 3$}
& \multicolumn{1}{c|}
{$\begin{bmatrix} {\rm conv}, 1\times 1, 64 \\ {\rm conv}, 3\times 3, 64 \\ {\rm conv}, 1\times 1, 256 \end{bmatrix} \times 3$}
&$N \times 16 \times 16 \times 256$     $N \times 64 \times 64 \times 256$  \\
\cline{2-4}
& \multicolumn{1}{c|}
{$\begin{bmatrix} {\rm conv}, 1\times 1, 128\\ {\rm conv}, 3\times 3, 128\\ {\rm conv}, 1\times 1, 512 \end{bmatrix}\times 4$} 
& \multicolumn{1}{c|}
{$\begin{bmatrix} {\rm conv}, 1\times 1, 128\\ {\rm conv}, 3\times 3, 128\\ {\rm conv}, 1\times 1, 512 \end{bmatrix}\times 4$} 
&$N \times 8 \times 8 \times 512$     $N \times 32 \times 32 \times 512$  \\
\cline{2-4}
& \multicolumn{1}{c|}{Global Maxpool}
& \multicolumn{1}{c|} {No Operation}
&$N \times 1 \times 1 \times 512$     $N \times 32 \times 32 \times 512$  \\
\cline{1-4}

\multirow{7}{*}{\emph{Matching}}
& \multicolumn{1}{c|}{Vector Broadcasting~($32 \times 32$)}
& \multicolumn{1}{c|} {No Operation}
&$N \times 32 \times 32 \times 512$     $N \times 32 \times 32 \times 512$  \\
\cline{2-4}
& \multicolumn{2}{c|}{Feature Map Concatenation}
&$N \times 32 \times 32 \times 1024$      \\
\cline{2-4}
& \multicolumn{2}{c|}{Relation \;Module}
& \\
& \multicolumn{2}{c|}
{$\begin{bmatrix} {\rm conv}, 3\times 3, 256\\ {\rm convt}, 3\times 3, 256 \end{bmatrix}$} 
&$N \times 64 \times 64 \times 256$      \\
\cline{2-4}
& \multicolumn{2}{c|}{Prediction}
& \\
& \multicolumn{2}{c|}
{$\begin{bmatrix} {\rm conv}, 3\times 3, 1 \end{bmatrix}$} 
&$N \times 64 \times 64 \times 1$      \\
\cline{2-4}

\hline
\end{tabular}}}
\end{center}
\caption{Architecture of the generic matching networks. 
``convt'' refers to convolutional transpose with stride 2.}
\label{net:tracknet}
\end{table}

\begin{figure}[H]
\begin{center}
\begin{tabular}{ccp{0.05cm}cc}
\includegraphics[width=0.215\textwidth]{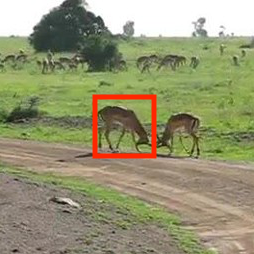} & 
\includegraphics[width=0.215\textwidth]{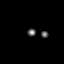} & &
\includegraphics[width=0.215\textwidth]{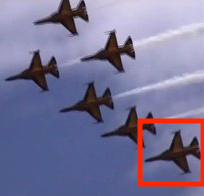} & 
\includegraphics[width=0.215\textwidth]{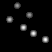} \\

\includegraphics[width=0.215\textwidth]{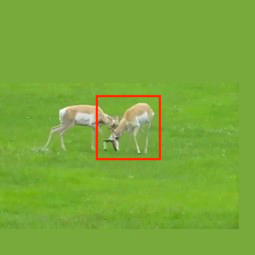} & 
\includegraphics[width=0.215\textwidth]{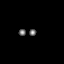} & &
\includegraphics[width=0.215\textwidth]{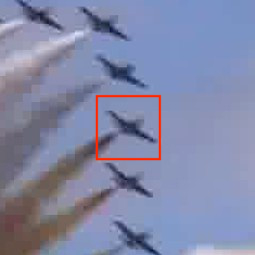} & 
\includegraphics[width=0.215\textwidth]{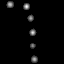} \\

\includegraphics[width=0.215\textwidth]{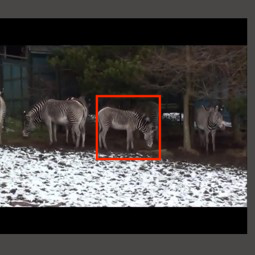} & 
\includegraphics[width=0.215\textwidth]{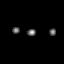} & &
\includegraphics[width=0.215\textwidth]{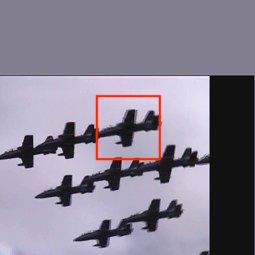} & 
\includegraphics[width=0.215\textwidth]{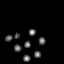} \\

\includegraphics[width=0.215\textwidth]{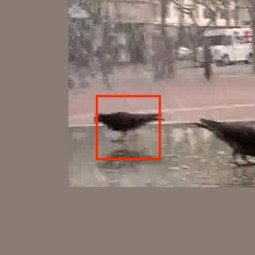} & 
\includegraphics[width=0.215\textwidth]{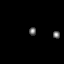} & &
\includegraphics[width=0.215\textwidth]{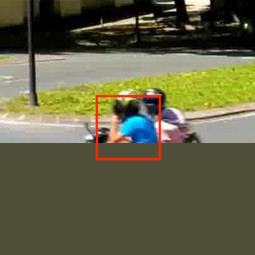} & 
\includegraphics[width=0.215\textwidth]{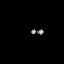} \\

\includegraphics[width=0.215\textwidth]{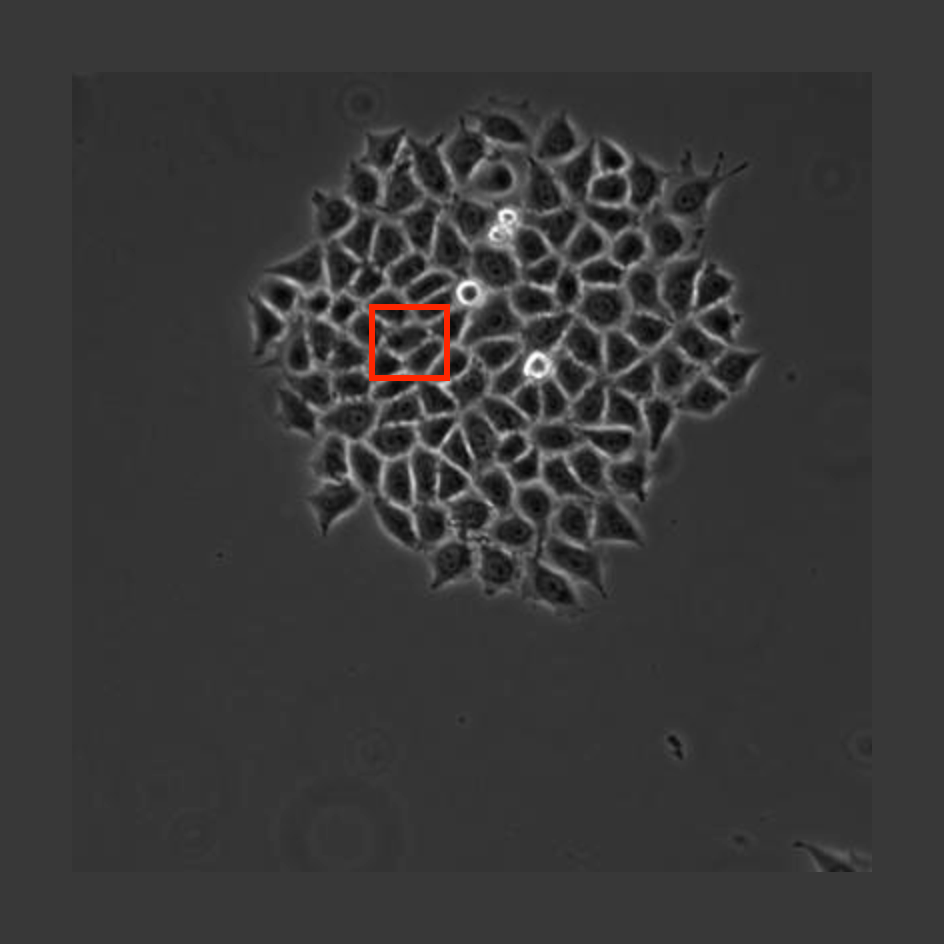} & 
\includegraphics[width=0.215\textwidth]{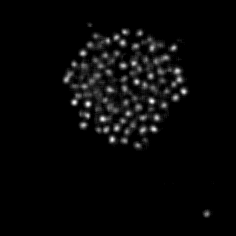} & & 
\includegraphics[width=0.215\textwidth]{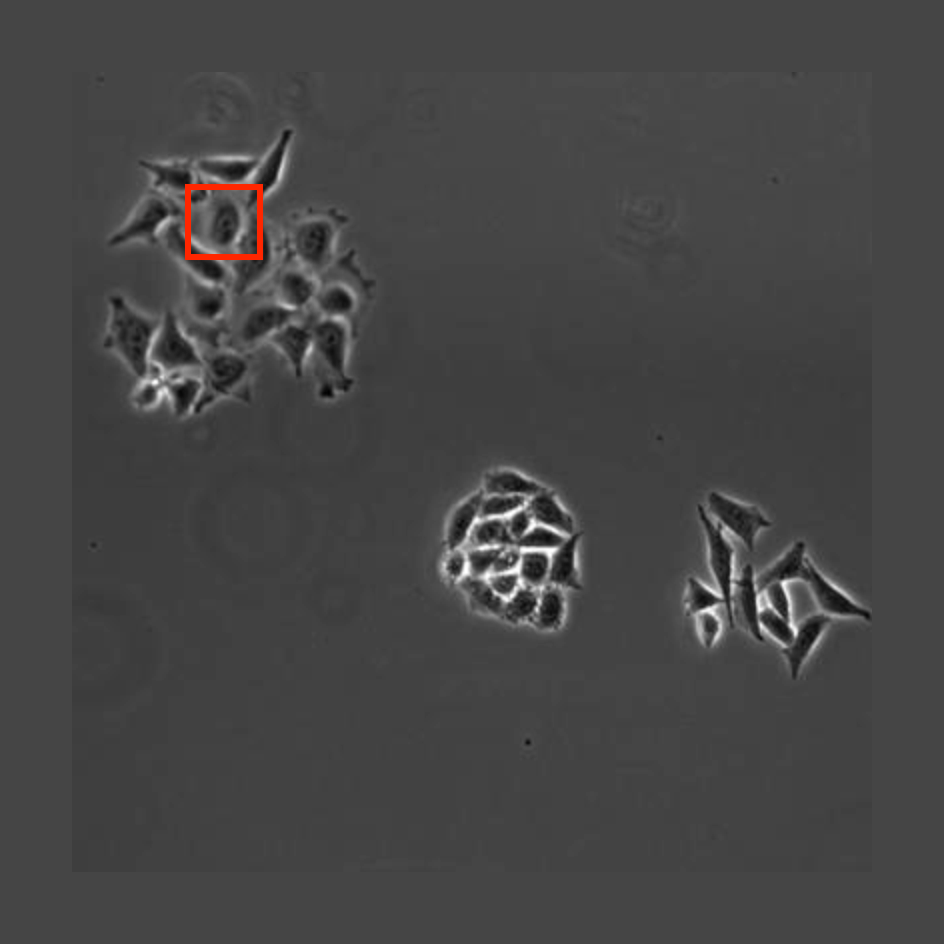} & 
\includegraphics[width=0.215\textwidth]{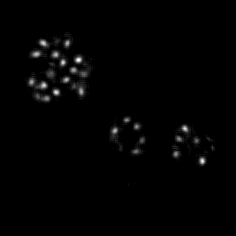} \\
Input & Prediction &  & Input & Prediction \\
\end{tabular}
\caption{Similarity predictions of the generic matching network on the video validation set, 
and on an \emph{unseen} dataset of HeLa cells.
The exemplar patch is marked with a red square.
Images are padded with the mean value and the resolution has been changed for visualization purposes.
As expected, the generic matching network has learned to match instances beyond a simplistic level;
for instance, the animals are of different viewpoints, the matched bird in the fourth row is partially occluded, 
and the people are in different colored shirts.
More interestingly, 
it acts as an excellent initialization for objects from unseen domains, 
even in the presence of large appearance and shape variation in the case of HeLa cells.}
\label{fig:tracking}
\end{center}
\end{figure}

Once trained on the tracking data, 
the model can be directly applied for detecting repetitions within an image. 
We show a number of  example predictions in Figure~\ref{fig:tracking}.
Note here, several interesting phenomena can be seen:
\emph{first}, 
as expected, the generic matching network has learned to match instances beyond a simplistic level;
for instance, the animals are of different viewpoints, 
the bird in the fourth row is partially occluded, 
and the persons are not only partially occluded, 
but also in different shirts with substantial appearance variations;
\emph{second}, object overlaps can also be handled, as shown in the airplane cases;
\emph{third}, although the ImageNet training set is only composed of natural images, 
and none of the categories has a similar appearance or distribution to  the HeLa cells, 
the generic matching network succeeds 
despite large appearance and shape variations which exist for cells.
These results validate  our idea of  building a class-agnostic counting network.
However, it is crucial to be able to easily {\em adapt}  the pre-trained model
to further specialize to new domains.

\subsection{Adapting}
The next objective is to specialize the network to new domains or new user requirements.
We add \textit{residual adapter modules}~\cite{Rebuffi18} implemented as $1\times1$ convolutions in parallel with the existing $3\times3$ convolutions in the embedding module of the network.
During adaptation, 
we freeze all of the parameters in the pre-trained generic matching network, 
and train only the adapters and batch normalization layers.
This results in 178K trainable parameters out of a total network size of 6.0M parameters, only 3\% of the total.

\subsection{Discussion and relation to prior work}
Object counting poses certain additional challenges that are less prominent or non-existent in tracking. 
First, 
rather than requiring a single maximum in a candidate window~(that localizes the object), 
counting requires a clean output map to distinguish multiple matches from noise and false positives. 
Second, unlike the continuous variation of object shape and appearance in the tracking problem,
object counting can have more challenging appearance changes, 
e.g.\ large degrees of rotation, and intra-class variation (in the case of cars, both color and shape). 
Thus, we find the approaches used in template matching~(SSD or cross-correlation~\cite{Bertinetto16,Dekel15,Leung96}) 
to be insufficient for our purposes~(as will be shown in Table~\ref{car results}).
To address these challenges, we learn a discriminative classifier $\gamma(\cdot; \theta_3)$
between the exemplar patch and search image, an idea that dates back to~\cite{Malisiewicz11}.

The residual adapters~\cite{Rebuffi18}  are added only to the embedding module, 
but we train the batch normalization layers throughout the entire network. 
Marsden~\etal~\cite{Marsden17} also use residual
adapters to adapt a network for counting different
objects.  However, they place the modules in the
final fully connected layers in order to regress a count, whereas we
add them to the convolutional layers in the residual blocks of the embedding module, 
such that they are able to change the filter responses at every stage of the base model,
providing more capacity for adaptation.

\section{Counting Benchmark Experiments}
\label{sec:exp}
As a proof of concept, 
the generic matching network is visually validated as a strong initialization for counting objects from unseen domains~(Figure~\ref{fig:tracking}).
To further demonstrate the effectiveness of the general-purpose GMN, 
we adapt the network to three different datasets: 
VGG synthetic cells~\cite{Lehmussola07,Lempitsky10b}, 
HeLa phase-contrast cells~\cite{Arteta15}, and
a large-scale drone-collected car dataset~\cite{Hsieh17}.

\begin{figure}[!htb]
\begin{center}
\setlength{\tabcolsep}{1pt}

\begin{tabular}{cp{0.01cm}c}
\includegraphics[width=0.49\textwidth]{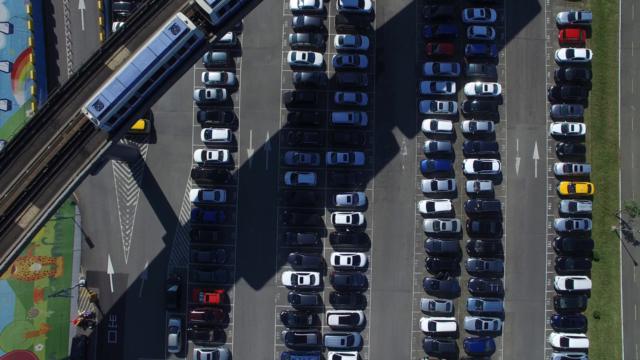} & 
&
\includegraphics[width=0.49\textwidth]{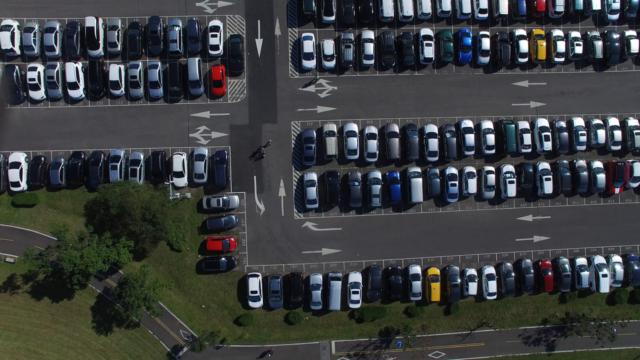} \\
Ground truth count: \textbf{117} & & Ground truth count: \textbf{132} \\
\includegraphics[width=0.49\textwidth]{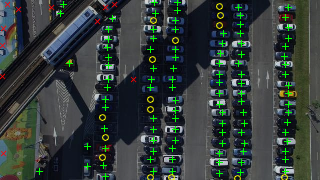} & 
&
\includegraphics[width=0.49\textwidth]{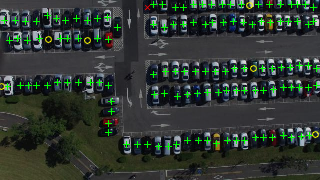} \\
Predicted count: \textbf{113}  & & Predicted count: \textbf{127}
\end{tabular}
\caption{Sample results on the CARPK dataset. Top row: original images. Bottom row: predicted detections. Correct detections are marked with a green `+', false positives with a red `x', and missed detections with a yellow `$\circ$'. Many of
the missed detections are dark cars in shadow, which upon inspection
are difficult for even a human eye to discern.}
\label{fig:cars}
\end{center}
\end{figure}

Each of these datasets poses unique challenges.
The synthetic cells contain many overlapping instances, 
a condition where density estimation methods have shown strong performance.
The HeLa cells exhibit significantly more variation in size and appearance than the synthetic cells,
and the number of training images is extremely limited~(only $11$ images); 
thus, detection-based methods with handcrafted features have shown good results.
In the car dataset, 
cars appear in various orientations, 
often within the same image, and can be partially occluded by trees and bridges; 
there is also clutter from motorbikes, buildings, and other distractors (Figure \ref{fig:cars}).
As shown in Hsieh~\etal~\cite{Hsieh17}, 
state-of-the-art models for object detection produce a very high error rate.
\subsection{Evaluation Metrics}
\label{subsec:eval}
The metrics we use for evaluation throughout this paper are the mean absolute counting error~(MAE), 
precision, recall, and F$_1$ score. 
To determine successful detections, 
we first take the local maximums (above a threshold $T$) of the predicted similarity map as the detections. 
$T$ is usually set as the value that maximizes the F$_1$ score on a validation set.
Note that, since multiple combinations of recall and precision can give the same F$_1$ score, 
we prioritize the recall score.
Following~\cite{Arteta15},
we then match these predicted detections with the ground truth locations using the Hungarian algorithm, 
with the constraint that a successful detection must lie no further than a tolerance $R$ from the ground truth location, 
where $R$ is set as the average radius of each object. 

\subsection{Synthetic fluorescence microscopy}
\begin{figure}[!htb]
\begin{center}
\setlength{\tabcolsep}{1pt}
\small
\begin{tabular}{cp{0.2cm}c}
\includegraphics[width=0.38\textwidth]{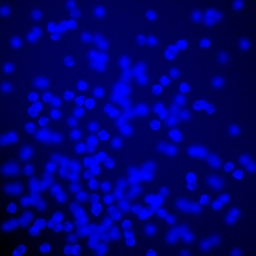} & &
\includegraphics[width=0.38\textwidth]{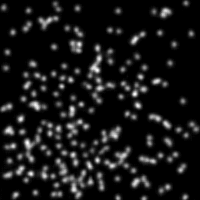}\\
True count: \textbf{254} & & Prediction: \textbf{252}  \\
\includegraphics[width=0.38\textwidth]{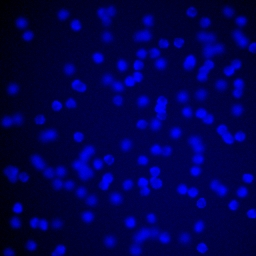} & &
\includegraphics[width=0.38\textwidth]{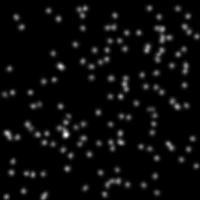} \\
True count: \textbf{142} & & Prediction: \textbf{142}
\end{tabular}
\caption{Example of counting results on synthetic cell images. For each pair of images, left: original image, and right: the network's predicted heat map, which is \emph{summed} to give the estimated count.}
\label{fig:syn_cell}
\end{center}
\end{figure}

The synthetic VGG cell dataset contains 200 fluorescence microscopy cell images, evenly split between training and testing sets.
We follow the procedure proposed by Lempitsky and Zisserman~\cite{Lempitsky10b} of sampling 5 random splits of the training set with $N$ training images and $N$ validation images. 
Results in Table~\ref{cell1_res} and Figure~\ref{fig:syn_cell} show that our method is not restricted to detection-based counting, 
but also performs well on density estimation-type problems in a setting with high instance overlap.
Note that, we compare with methods that are highly engineered for this dataset.
\begin{table}[!htb]
\begin{center}
\begin{tabular}{|l|c|c|c|c|}
\hline
Method & MAE & Precision & Recall & F$_1$-score \\
\hline\hline
Xie \etal \cite{Xie15} & 2.9 $\pm$ 0.2 & - & - & - \\
Fiaschi \etal \cite{Fiaschi12} & 3.2 $\pm$ 0.1 & - & - & - \\
Lempitsky and Zisserman \cite{Lempitsky10b} & 3.5 $\pm$ 0.2 & - & - & - \\
Barinova \etal \cite{Barinova10} & 6.0 $\pm$ 0.5 & - & - & - \\
Singletons \cite{Arteta15} & 51.2 $\pm$ 0.8 & 98.87 $\pm$ 1.52 & 72.07 $\pm$ 0.85 & 83.37 $\pm$ 1.20\\
Full system w/o surface \cite{Arteta15} & 5.06 $\pm$ 0.2 & 95.00 $\pm$ 0.75 & 91.97 $\pm$ 0.43 & 93.46 $\pm$ 0.15\\
Ours & 3.56 $\pm$ 0.27 & 99.43 $\pm$ 0.05 & 82.50 $\pm$ 0.15 & 90.18 $\pm$ 0.07 \\
\hline
\end{tabular}
\end{center}
\caption{Results for the synthetic cell dataset. 
All methods are trained on the $N=32$ split. 
Standard deviations are calculated using 5 random splits of training and validation sets and 5 randomly sampled exemplar patches per image.
Note here, the exemplar patches are sampled from images in the training set,
and different exemplar patches have negligible effect on performance.}
\label{cell1_res}
\end{table}

\subsection{HeLa cells on phase contrast microscopy}
\label{sec:hela_exp}
\begin{figure}[!htb]
\begin{center}
\setlength{\tabcolsep}{1pt}
\begin{tabular}{cp{0.2cm}c}
\includegraphics[width=0.38\textwidth]{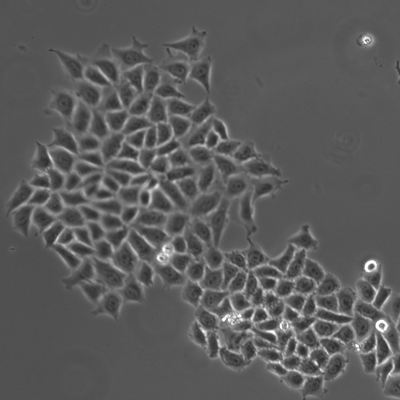} & &
\includegraphics[width=0.38\textwidth]{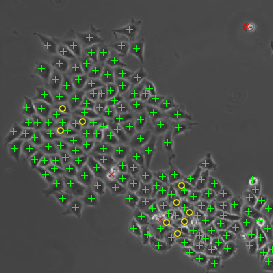} \\
True count: \textbf{177} & &  Prediction: \textbf{171} \\
& & Matches: 168 \\
\includegraphics[width=0.38\textwidth]{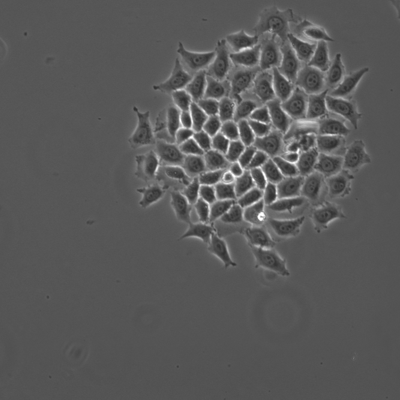} & &
\includegraphics[width=0.38\textwidth]{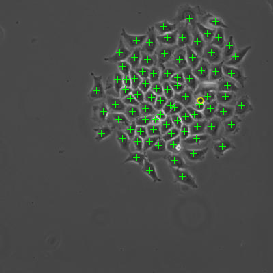} \\
True count: \textbf{85} & & Prediction: \textbf{84}\\
& & Matches: 84 \\
\end{tabular}
\caption{Example detection results on the HeLa cell test set. 
Correct detections~(based on Hungarian matching) are marked with a green `+', false positives with a red `x', 
and missed detections with a yellow `$\circ$'.}
\label{fig:hela_cell}
\end{center}
\end{figure}

The dataset contains 11 training and 11 testing images. 
We follow the training procedure of~\cite{Arteta15} and train in a leave-one-out fashion for selecting hyperparameters, 
e.g.\ detection threshold $T$. 
Results are shown in Table~\ref{cell2_res}. 
As shown in Figure~\ref{fig:hela_cell}, 
our method performs well in scenarios of large intra-class variations in shape and size, 
where SSD and cross-correlation would suffer.
Overall, 
our GMN  achieves  comparable results to the conventional methods with hand-crafted features,
despite the training dataset being extremely small for current deep learning standards.

\begin{table}[!htb]
\begin{center}
\begin{tabular}{|l|c|c|c|c|}
\hline
Method & MAE & Precision & Recall & F$_1$-score \\
\hline\hline
Correlation clustering \cite{Zhang14} & - & - & - & 95 \\
Singletons \cite{Arteta15} & 2.36 $\pm$ 0.67 & 93.70 $\pm$ 0.20 & 91.94 $\pm$ 0.72 & 92.81 $\pm$ 0.35\\
Full system w/o surface \cite{Arteta15} & 3.84 $\pm$ 1.44 & 98.51 $\pm$ 1.16 & 95.76 $\pm$ 0.27 & 97.10 $\pm$ 0.27\\
Ours & 3.53 $\pm$ 0.18 & 96.05 $\pm$ 0.04 & 94.22 $\pm$ 0.06 & 95.12 $\pm$ 0.05\\
\hline
\end{tabular}
\end{center}
\caption{Results for the HeLa cell dataset. We calculate MAE using the detection counts,
since the instances are well-separated.
Our standard deviations are calculated using 5 randomly sampled exemplar patches per image.
Note here, the 5 exemplar patches are sampled from images in the training set;
different exemplar patches have negligible effect on performance.}
\label{cell2_res}
\end{table}

\subsection{Cars}
We next demonstrate the GMN's performance on counting cars in aerial images. 
This drone-collected dataset~(CARPK) consists of 989 training images and 459 testing images~(nearly 90,000 instances of cars), 
where the images are taken from overhead shots of car parking lots.
The training images are taken from three different parking lot scenes, 
and the test set is taken from a fourth scene. 
We compare our network to the region proposal and classification methods in Table~\ref{car results}.

In the experiments, we train two GMN models with augmentation: 
one on just three  images (99 total cars) randomly sampled from the training scenes, 
which achieves state-of-the-art results, and
one on the full CARPK training set, which further boosts the performance by a large margin. 

\begin{table}[!htb]
\begin{center}
\begin{tabular}{|l|c|c|c|c|c|}
\hline
Method & $T$ & MAE & RMSE & Recall & Precision\\
\hline\hline
*YOLO~\cite{Hsieh17,Redmon16} & - & 48.89 & 57.55 & - & -\\
*Faster R-CNN~\cite{Hsieh17,Ren16} & - & 47.45 & 57.39 & - & -\\
*Faster R-CNN (RPN-small)~\cite{Hsieh17,Ren16}& - & 24.32 & 37.62 & - & -\\
$\dagger$One-Look Regression~\cite{Hsieh17,Mundhenk14} & - & 59.46 & 66.84 & - & -\\
*Spatially Regularized RPN~\cite{Hsieh17} & - & 23.80 & 36.79 & 57.5\% & -\\
\hline
Template matching (Sum of Squared Distances) &- & 49.8 & 59.7 & 20.0\% & 29.1\%\\
Ours (3 images, 99 cars) & 2.5 & 36.71 & 44.16 & 60.65\% & 93.91\% \\
Ours (3 images, 99 cars) & 2 & 22.32 & 28.72 & 71.32\% & 90.23\% \\
Ours (3 images, 99 cars) & 1.75 & 17.32 & 22.81 & 74.16\% & 87.87\% \\
Ours (3 images, 99 cars) & 1.5 & 13.38 & 18.03 & 76.1\% & 85.1\% \\
Ours (full dataset) & 2.75 & 19.66 & 25.12 & 78.61\% & 97.0\%\\
Ours (full dataset) & 2.5 & 14.36 & 19.01 & 83.2\% & 96.0\% \\
Ours (full dataset) & 2 & 8.38 & 11.55 & 87.46\% & 93.4\%\\
Ours (full dataset) & 1.75 & \textbf{7.48} & \textbf{9.9} & 88.4\% & 91.8\%\\
\hline
\end{tabular}
\end{center}
\caption{Mean Absolute Error (MAE), Root Mean Squared Error (RMSE), 
Recall and Precision comparisons on the CARPK dataset. 
The ``*'' indicates that the method has been fine tuned on the full dataset, 
and the ``$\dagger$'' indicates that the method has been revised to fit the dataset, 
as described in \cite{Hsieh17}. 
We show our method trained on 3 images and on the full dataset, 
with varying thresholds $T$. 
We calculate MAE using 5 randomly sampled exemplar patches per image,
and the final counts are obtained from local maximums~(counting by detection).
Note here, the exemplar patches are sampled from images in the training set,
and different exemplar patches have negligible effect on performance.
Standard deviation is not reported in this table, 
but can be easily computed
following the previously reported manner.}
\label{car results}
\end{table}

When determining counts based on local maximums, 
we note it is possible that our model outperforms the previous detection-based methods due to 
false positives and false negatives ``canceling'' each other, making the counting error very low.
Thus, 
we investigate effects of the threshold $T$~(as defined in \S~\ref{subsec:eval}) on selecting detections from candidate local maximums, and report results for several values of $T$.
Note that, by varying this hyperparameter, 
we are able to explicitly control the precision-recall of our model.
While calculating recall and precision, 
we consider a detection to be successful if it lies within 20 pixels~(determined based on the mean car size) of the ground truth location. 
The recall reported for the region proposal methods in Table~\ref{car results} is calculated by averaging across scores from using various IoU thresholds, as described in~\cite{Hsieh17}.

As shown in Table~\ref{car results}, 
the MAE is calculated  with 5 randomly sampled exemplar patches per image,
and the final counts are obtained by counting local maximums~(detection-based counting).
Note here, the exemplar patches are sampled from images in the training set,
and different exemplar patches have negligible effect on performance.
We can see that even with a very high precision~(model trained on the full dataset, with $T=2.75$), 
our model can still outperform the previous state-of-the-art by a substantial margin~(counting error: MAE=$23.8$ vs MAE=$19.7$).
Further decreasing the threshold yields higher recall at the expense of precision, with
our best model achieving a counting error of MAE=$7.5$.

\subsection{Discussion}
From our experiments, the following phenomena can be observed:

\noindent
\emph{First}, in contrast to previous work, 
where different architectures are designed for density estimation in scenarios with significant instance overlap (e.g.\ VGG synthetic cells) 
and for detection-based counting in scenarios with well-separated objects~(e.g.\ HeLa cells and cars),
the GMN  has the flexibility to handle both scenarios. 
Based on the amount of instance overlap, the object counts can simply be obtained by taking either the integral
in the former case, or the local maximum in the latter, or possibly even an ensemble of them~\cite{Idrees18}.

\noindent
\emph{Second}, 
by training in a discriminative manner,
the GMN is able to match instances beyond the simplistic level, 
making it more robust to large degrees of rotation 
and appearance variation than the baseline SSD-based template matching~(as shown in Table~\ref{car results}). 

\noindent
\emph{Third}, 
in the cases where training data is limited~(11 images for HeLa cells, 3 for cars),
the proposed model has consistently shown comparable or superior performance to the state-of-the-art methods, 
indicating the model's ease of adaptation, 
as well as verifying our observation that videos can be a natural data source for learning \emph{self-similarity}.

\section{Shanghaitech Crowd Counting}
\label{sec:exp2}

To further demonstrate the power and flexibility of counting-by-matching,
we extend it to the Shanghaitech crowd counting dataset, 
which contains images of very large crowds of people from arbitrary camera perspectives,
with individuals appearing at extremely varied scales due to perspective.

We carry out a preliminary implementation of our method on the Shanghaitech Part A crowd dataset. 
Inspired by the idea of crowd detection as repetitive textures~\cite{Arandjelovic08},
we conjecture that it is possible to ignore individual instances
and match the statistics of patches instead; 
e.g.\ the statistics of patches with 10 people should be different from those with 20 people.

\begin{figure}[!htb]
\begin{center}
\setlength{\tabcolsep}{1pt}
\small
\includegraphics[width=\textwidth]{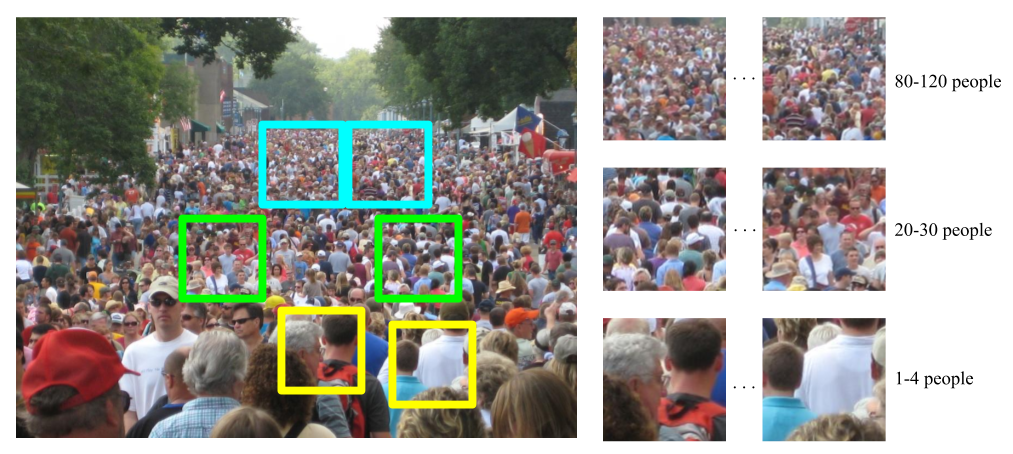}
\caption{Example of how different patches are classified.
Patches marked by the same color square belong to the same density class.
As can be seen, the significant textural differences between the classes
enables a network to learn to classify different densities.}
\label{fig:crowd}
\end{center}
\end{figure}

We take the following steps: 
(1) Using the ground truth dot annotations, 
we quantize $64\times64$ pixel patches into 10 different classes based on number of people, 
e.g. one class will be 0 people, another 5 people, etc. (See Figure \ref{fig:crowd} for an example.)
(2) Following the idea of counting-by-matching, 
we train the self-similarity architecture to embed the patches based on the number of people,
i.e.\  if patches are sampled from the same class, the model must predict 1, otherwise 0.
(3) We run the model on the test set using a sample of each class from the training set as the exemplar patch, 
with the final classification made by the maximum response.

Compared to other models that are specifically designed to count human crowds 
(e.g.\  CNNs with multiple branches),
we aim for a method with the potential for low-shot category-agnostic counting. 
Our preliminary experiments show the possibility of scaling the counting-by-matching idea to human crowd datasets. 
\begin{table}[!htb]
\begin{center}
\begin{tabular}{|l|c|c|}
\hline
Method & MAE & RMSE\\
\hline\hline
$\dagger$Zhang \etal~\cite{Zhang15a} & 181.8 & 277.7\\
MCNN-CCR~\cite{Zhang16} & 245.0 & 336.1\\
MCNN~\cite{Zhang16} & 110.2 & 173.2\\
ic-CNN~\cite{Ranjan18} & \textbf{69.8} & \textbf{117.3} \\
Ours & 95.8 & 133.3\\
\hline
\end{tabular}
\end{center}
\caption{
Preliminary results on Shanghaitech Part A (lower is better). Patches chosen based on validation set performance.
The ``$\dagger$'' result is from paper~\cite{Zhang16}.}
\label{tab:crowd}
\end{table}
\section{Conclusion}
In this work, we recast counting as a matching problem, which offers
several advantages over traditional counting methods. Namely, we make
use of object detection video data that has not yet been utilized by
the counting community, and we create a model that can flexibly adapt
to various domains, which is a form of few-shot learning.
We hope this unconventional structuring of the
counting problem encourages further work towards an all-purpose
counting model.

Several extensions are possible for future works:
\emph{first}, it would be interesting to consider counting in video sequences, 
rather than individual images or frames. 
Here the tracking analogue takes on an even greater significance as a counting model
can take advantage of both within-frame and between-frame similarities,
\emph{second}, a carefully engineered scale-invariant network with more sophisticated feature fusion than the 
GMN  could potentially improve the current results.
\section*{Acknowledgements}
Funding for this research is provided by the Oxford-Google DeepMind Graduate Scholarship, 
and by the EPSRC Programme Grant Seebibyte EP/M013774/1.

 \bibliographystyle{splncs04}
\bibliography{shortstrings,vgg_local,vgg_other}
\end{document}